\RequirePackage{fix-cm}
\documentclass[smallextended]{svjour3}       % onecolumn (second format)
\smartqed  % flush right qed marks, e.g. at end of proof
\usepackage{graphicx}
\usepackage{amsmath}
\usepackage{amsfonts}
\usepackage{amssymb}
\usepackage{multirow}
\usepackage{rotating}
\usepackage{makecell}
\usepackage{lineno}
\begin{document}
	
	\title{Identifying Hazardousness of Sewer-Pipeline Gas-Mixture using Classification Methods%\thanks{Grants or other notes
	}
	\subtitle{A Comparative Study}
	
	\titlerunning{Identifying Hazardousness of Sewer-Pipeline}        % if too long for running head
	
	\author{Varun Kumar Ojha \and Parmartha Dutta \and Atal Chaudhuri
		%etc.
	}
	
	
	\institute{V. K. Ojha \at
		\emph{IT4Innovations, V{\v{S}}B Technical University of Ostrava, Ostrava, Czech Republic} and Dept. of Computer Science \& Engineering, Jadavpur University, Kolkata, India \\
		\email{varun.kumar.ojha@vsb.cz}           %  \\
		\and
		P. Dutta \at
		Dept. of Computer \& System Sciences, Visva-Bharati University, India\\
		\email{paramartha.dutta@gmail.com}
		\and
		A Chaudhuri \at
		Dept. of Computer Science \& Engineering, Jadavpur University, Kolkata, India
		\email{atalc23@gmail.com}
		~\\
		~\\Neural Computing and Applications~\\
		DOI: 10.1007/s00521-016-2443-0
	}
	
	\date{Received: date / Accepted: date}
	
	\maketitle
	
	\begin{abstract}
		In this work, we formulated a real-world problem related to sewer-pipeline gas detection using the classification-based approaches. The primary goal of this work was to identify the hazardousness of sewer-pipeline to offer safe and non-hazardous access to sewer-pipeline workers so that the human fatalities, which occurs due to the toxic exposure of sewer gas components, can be avoided. The dataset acquired through laboratory tests, experiments, and various literature-sources were organized to design a predictive model that was able to identify/classify hazardous and non-hazardous situation of sewer-pipeline. To design such prediction model, several classification algorithms were used and their performances were evaluated and compared, both empirically and statistically, over the collected dataset. In addition, the performances of several ensemble methods were analyzed to understand the extent of improvement offered by these methods. The result of this comprehensive study showed that the instance-based-learning algorithm  performed better than many other algorithms such as multi-layer perceptron, radial basis function network, support vector machine, reduced pruning tree, etc. Similarly, it was observed that multi-scheme ensemble approach enhanced the performance of base predictors.
		
	\end{abstract}
	
	\section{Introduction}
	\label{intro}
	This is in the view of providing a solution to a real-world problem using technology, where the human fatalities need to be avoided. Hence, the technology should be as simple as possible. In this work, we addressed a complex real-world problem related to sewer-pipeline gas detection, where sewer-pipeline safety detection (in terms of non-toxic environment) was required to allow maintenance and cleaning of the pipeline. The sewer gas detection is a highly complex problem because of the presence of several toxic gases in a mixture form, and a single gas detector may not offer reliable solution. Therefore, we studied the complexity of this problem in terms of gas mixture. The primary goal was to offer a simple solution with a high accuracy so that it was easy to categorize the hazardous situation in straightforward way such as ``hazardous" or ``non-hazardous." To meet this simplicity, we formulated sewer-pipeline gas detection problem as a classification problem. 
	
	Sewer-pipeline contains a mixture of several toxic gases such as hydrogen sulphide (H$_2$S), ammonia (NH$_3$), methane (CH$_4$), carbon dioxide (CO$_2$), nitrogen oxides (NO$_x$), etc.,~\cite{SewerGas,lewis1996material,SewerGas1}. Usually, this mixture is generated due to the biodegradation of the waste and the sewage into the sewer-pipeline. Such toxic gas-mixture is fatal for those who come to the proximity/exposure of these gases. Following this, an alarming number of human fatalities are reported each year by the newspapers and the other agencies~\cite{Hindu2014a,NDTV2015,Tehelka2007}. The authorities those are responsible for maintaining and cleaning of the sewer pipeline provides various electronic portable gas detectors available in the market to the employed persons so that they can determine the safeness of the sewer-environment before physically get involve into the maintenance work. However, the available electronic portable gas detectors are not providing satisfactory results. It is evident from the recent comments from the judiciary to these authorities. In a judgment to a civil appeal number 5322 of 2011, the Supreme Court of India stated, ``the State and its agencies/instrumentalities cannot absolve themselves of the responsibility to put in place effective mechanism for ensuring safety of the workers employed for maintaining and cleaning the sewage system~\cite{Hindu2014d}." Similarly, in another judgment, the Supreme Court of India stated, ``...entering sewer lines without safety gears should be made a crime even in emergency situations...~\cite{Hindu2014b,Hindu2014c}." This motivated us to carry out our research in this domain and to come out with a simple solution so that without having the minimum knowledge of the technicalities of gas composition and safety limits, a person is able to understand the environment of a sewer system before entering.
	
	To ensure the simplicity in model, we collected and preprocessed data to realize sewer gas-detection as a binary class classification problem. However, in this work, apart from the objective of constructing a prediction model, we set a secondary objective, which was to analyze the performances of the classifiers, both empirically and statistically. To meet these objectives, we used 12 base predictors from four different categories such as neural network based classifiers, tree based classifiers, instance based classifiers, and rule based classifiers. The algorithms were applied over the collected dataset and the performance of the algorithms were collected in terms of the accuracy. The collected results were then used for analyzing the performance superiority of the one algorithm over another or the one category of algorithms over another. 
	
	We observed that the performance of the algorithms were independent of the category they belong to. For example, the performance of instance based k-nearest neighbor, logistic model tree, and support vector machine came from three different categories, but they had a very competitive performance. However, we must consider the ``No-free-lunch theorem" that suggests that some algorithms perform better on some problem and some on another~\cite{wolpert1997no}. Therefore, to find out which predictor performs best in this case, we used 12 base predictors and nine ensemble methods.
	
	Rest of the article is organized as follows. A background study is provided in Section~\ref{sec:2.1}, which leads to setting ground for describing our contribution to the sewer-pipeline gas detection. In Section~\ref{sec:2.2}, we provide a detailed description of the data collection and preprocessing mechanisms, which constitute the core and significant part for formulating gas detection problem as a binary classification problem. Section~\ref{sec:2.3} deals with the brief descriptions of the classifiers/algorithms and methods used for constructing the prediction model. The design of comprehensive experiment set for the evaluation of the classifiers is reported in Section~\ref{sec:3}. Whereas, Section~\ref{sec:3} describes empirical and statistical evaluation of the classifiers, discussions and conclusion are reported in Sections~\ref{sec:4} and~\ref{sec:5}, respectively.
	\section{Methodology}
	\label{sec:2}
	In this Section, we put together the background study, the data collection mechanisms, and the classification methods definitions. The background study describes the significance of the sewer-pipeline gas detection problem and the data collection mechanism describes the formulation of gas-detection as a classification problem.
	
	\subsection{Background Study}
	\label{sec:2.1}
	Literature review was conducted in the perspective of electronic-nose (E-NOSE) and gas-detection-system to cover a broad area of research in the field of gas detection and modeling using intelligent computing techniques/algorithms. Although not much work specifically on sewer gas-mixture-detection was reported in the past, few notable contributions were observed. Li et al.~\cite{jun1993enose} reported a noticeable research work on the development and design of an electronic nose (E-NOSE) and gas detection system, where a neural network (NN)-based mixed gas (NOx, and CO) measurement system was developed. On the other hand, Sirvastava et al.~\cite{sirvastava2000enose,sirvastava2000enoseGA} proposed a design of intelligent E-NOSE system using backpropagation (BP) and neuro-genetic approach. Llobet et al.~\cite{eduard2001enose} presented a pattern recognition approach, based on the wallet transformation for gas mixture analysis using single tin-oxide sensor. Liu et~al.~[13] addressed a genetic-NN algorithm to recognize patterns of mixed gases (a mixture of three component gases) using infrared gas sensor. Lee et al.~\cite{dae2005enose} illustrated uses of micro gas sensor array (GSA) combined with NN for recognizing combustible leakage gases. Ambard et al.~\cite{maxim2008enose} have demonstrated use of NN for gas discrimination using a tin-oxide GSA for the gases H$_2$, CO and CH$_4$. In~\cite{hakim2009enose}, authors have illustrated a NN-based technique for developing a gas sensory system for sensing gases in a dynamic environment. Pan et al.~\cite{wu2009enose} have shown several applications of E-NOSE. Wongchoosuka et al.~\cite{chatchawal2010enose} have proposed an E-NOSE detection system based on carbon nanotube-$ SnO_2 $ gas sensors for detecting methanol. Zhang et al.~\cite{qian2010enose} developed a knowledge-based genetic algorithm for detecting mixed gas in mines. Won et al.~\cite{won2010enose} proposed a system for estimation of hazardous gas release rate using optical sensor and NN-based technique. The following salient points came out of  the above mentioned articles:
	\begin{itemize}
		\item Mainly, BP and NN-based approaches were studied so far for detecting gas-mixtures.
		\item Mostly, the E-Nose systems reported in the past were developed for the gas-mixtures of only two or three gases and the sensors of the gases used were less cross-sensitive to the other gases in mixtures.
		\item Cross-sensitivity during sensing is an important factor in gas detection system, which was least reported in literature as yet. However, Ojha et al.~\cite{Ojha_ga_j,Ojha_pso_j,Ojha_ch_bp,Ojha_ch_cg,ojha2016understating,ojha2016multi} offered a few methods such as neuro-genetic, neuro-swarm, ant-colony-based, neuro-simulated annealing, etc., where  cross-sensitivity factor has been addressed to some extent. However, these works were primarily related to regression modeling.
		\item The impact of humidity and temperature on sensors remained ignored so far. 
		\item The gas detection system or E-Nose was viewed only in the framework of regression problems and not classification problem. 	
	\end{itemize}
	
	Classification based approach led us to determine the hazardous and non-hazardous situation of a sewer-pipeline. In addition, the collection, organization, and the preprocessing of the collected data enabled us to address the cross-sensitivity issue firmly. The cross-sensitivity issue occurs because of the sensitivity of one gas-sensor towards multiple gases. So was our case, where a semiconductor-based GSA was designed using five gas-sensors. Each gas-sensor was typically meant for detecting its respective target gas. Hence, when the GSA was used for collecting data for a mixture of gases, the cross-sensitivity in the sensed values (collected data) became inevitable. Therefore, rather than considering pure results of the respective gases, we registered the cross-sensitive results as a part of our-collected data. Since a computationally intelligent model learned from the data and  also maintained the cross-sensitivity patterns registered in terms of data values itself, a  learned model accurately predicts an unknown gas mixture.
	
	\subsection{Equipment and Data Collection Mechanism}
	\label{sec:2.2}
\begin{figure}[h!]
		\centering
		\includegraphics[scale=0.8]{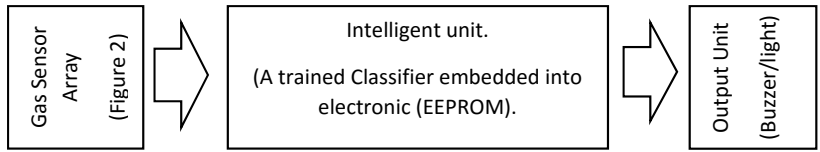}
		\caption{Block diagram of intelligent system design (real time data flow process) }
		\label{fig:1}       % Give a unique label
	\end{figure}

	Before explaining the details of data collection and equipment, we need to explain the basic design and the purpose of our work, which is to offer an intelligent gas detection system (an electronic portable gas detector) that will be a result of embedding learned-predictor (trained-classifier) into an electronic system. The data flow into our developed intelligent system is shown in Fig.~\ref{fig:1}, which describes the entire process of the intelligent system design, which is divided into three phases: 1) The data acquisition unit, which consists of gas suction-motor chamber, GSA, and data acquisition-cum data-preprocessor block; 2) An intelligent unit (classifier unit), which receives data from data-acquisition unit and classifying the acquired data patterns; 3) The output unit, which prompts the result in terms of colored light and buzzer. Hence, our objective here was limited to only train a classifier using the collected data. We describe the data collection process as follows.
	
	%\paragraph{Paragraph headings} Use paragraph headings as needed.
	%\begin{equation}
	%\end{equation}
	
\begin{figure}[h!]
		\includegraphics[scale=0.3]{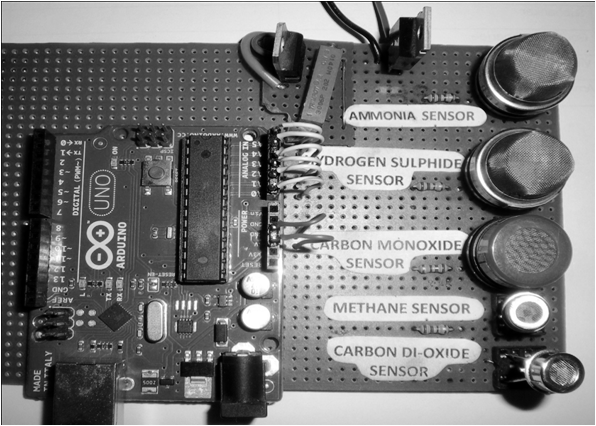}
		\includegraphics[scale=0.3]{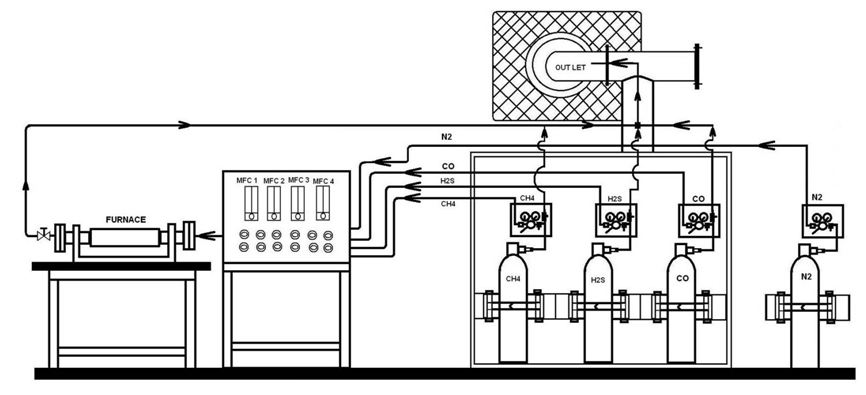}
		\caption{Laboratory-scale gas sensor array (GSA)~\cite{GSA,PortableGSA}}
		\label{fig:2}       % Give a unique label
	\end{figure}

	At first, we collected the data samples from the data-sheets, literature, and laboratories test of the collected gas mixture samples from sewer-pipelines. Second, we designed our own metal oxide semiconductor (MOS) gas sensors array (GSA) that was used for verifying the literature and laboratory data and for generating the data samples for the purpose experiments. Our designed GSA consists of five gas-sensors for sensing five different gases. They include hydrogen sulphide (H$_2$S), ammonia (NH$_3$), methane (CH$_4$), carbon dioxide (CO$_2$), and nitrogen oxides (NO$_x$). Typically, MOS sensors are resistance-type electrical sensors, where responses are change in circuit resistance proportional to gas concentration, A resistance type sensor responds to change in resistance due to change in the concentration of gases. The change in resistance is given as $\delta R_s/R_0$, where $\delta R_s$ is change in MOS sensor resistance and $R_0$ is base resistance or the sensing resistance at a specifics gas concentration in clean air~\cite{chatchawal2010enose}. The $R_0$ of the sensors MiCS - 4514, MQ - 7, MQ - 136, MQ - 135, and MQ - 4 is 0.25 ppm, 100 ppm, 10 ppm, 100 ppm and 1000 ppm, respectively. Here, ppm is the unit for measuring concentration of gas into air which is defined as follows: 1 ppm is equal to $ 1 $ volume of a gas into $ 10^6 $ volume of air. A typical arrangement of a gas sensor array is shown in Fig.~\ref{fig:2}. The circuitry shown in Fig.~\ref{fig:2} (left) was developed in our laboratory. Here, the fabricated and installed sensors were MiCS - 4514, MQ - 7, MQ - 136, MQ - 135, and MQ - 4 for gases NO$_2$, CO, H$_2$S, NH$_3$, and CH$_4$, respectively~\cite{GSA,PortableGSA}.
	
	The gas sensors used were sensitive to not only their target gases, but they were sensitive also to other gases in the gas-mixture~\cite{cantalini2003sensitivity,mitzner2003development}. Hence, cross-sensitivity effect over MOS sensors was confirmed~\cite{junhua2001enose}. It was moreover confirmed that the sensor responses were noisy and accordingly the pattern of such noise were considered and recorded as an instance into our dataset. Hence, a non-intelligent use of raw values of sensor response for hazardousness prediction may be misleading in operating (real-world) environment. Therefore, a training electronic portable gas detector  may be used to predict sewer hazardousness, accurately. So was the effort in this work to provide a classifier.
	
	Data collection had vital role in training of a classifier. Data samples were collected as per the following steps. At first, several manhole samples collected from the Kolkata, India municipal area were tested in laboratory to identify the presence of several toxic gases such as nitrogen dioxide (NO$_2$), carbon monoxide (CO), hydrogen sulphide (H$_2$S), ammonia (NH$_3$), methane (CH$_4$), and carbon dioxide (CO$_2$). Secondly, gas sensors were identified for each of the respective gases. As a result we came out with the procurement of gas sensor MiCS - 4514, MQ - 7, MQ - 136, MQ - 135, and MQ - 4 for NO$_2$, CO, H$_2$S, NH$_3$, and CH$_4$, respectively. We collected data sheets form the companies for the respective sensors. In the third step, a laboratory was setup for the verification and collection of the sensor response of the respective gas sensors  in certain range of their concentration. Specifically, the concentration range in ppm laid down in sensor manuals of sensors MiCS - 4514, MQ - 7, MQ - 136, MQ - 135, and MQ - 4 are [0.25 - 5], [20 - 1000], [1 - 100], [10 - 300] and [300 - 10000] of the gases NO$_2$, CO, H$_2$S, NH$_3$, and CH$_4$, respectively. In addition, the lab was setup (see Fig.~\ref{fig:2} [right]), where gas cylinders were connected to a gas concentration measuring unit called mass flow controller (MFC), which was further connected to a gas chamber, where each gas was allowed to pass in a specific concentration over an array of gas sensor. More specifically, the behavior of each of the gas sensors was recorded.
	
	%\begin{figure*}
	%  \includegraphics[width=0.75\textwidth]{example.eps}
	%\end{figure*}
	%
	The following steps were used for preparing data sample for the classifiers' training. First, hazardous (safety) limits of the component gases of manhole gas mixture were collected. Secondly, three different levels, (i) above safety-limit, (ii) at safety-limit, and (iii) below safety-limit for each manhole gas were recognized. Thirdly, gases were mixed in different combination to prepare several mixture sample that were used to pass over GSA. Table~\ref{tab:combination} indicates few examples of such mixture of gases in different combinations. For example, when we mix five gases each of which has three different recognized concentration levels, we get 243 different combinations  ($3^5$). In addition, we considered the role of humidity and temperature to influence the sensor's behavior. Accordingly, the data values were recorded. Hence, our collected dataset contained seven input features and an output class. Each sample was labeled with ``0" for safe sample (if the responses of all five sensors were under the maximum safety limit) or ``1" for unsafe sample (if the responses of any among the five sensors were above the maximum safety limit). The safety limits of the manhole gases are as follows: safety limit of NH$_{3}$ is between 25 ppm and 40 ppm~\cite{donham2002exposure}, CO is in between 35 ppm and 100 ppm~\cite{gasCO1998}, H$_{2}$S is in between 50 ppm and 100 ppm~\cite{gasH2S2007}, CO$_{2}$ is in between 5000 ppm and 8000 ppm~\cite{gasCO22007} and CH$_{4}$ is in between 5000 ppm and 10000 ppm~\cite{gasCH42002}. Table~\ref{tab:1} illustrates a fraction of the collected data samples.
	\begin{table}[b!]
		\caption{Samples of gas-mixture in different concentration.}
		\label{tab:combination}       % Give a unique label
		\begin{tabular}{llllllllll}
			\hline\noalign{\smallskip}
			& & & \multicolumn{5}{c}{Concentration of gases in ppm} & & \\
			\cline{4-8}\noalign{\smallskip}
			\# &	Humidity &	Temperature &	NO2 &	CO &	H2S &	NH &	CH &	Class & Status \\
			\noalign{\smallskip}\hline\noalign{\smallskip}
			1 & 65 & 20 & 0 & 10 & 10 & 20 & 2000 & 0 & safe\\
			2 & 65 & 20 & 0 & 10 & 10 & 20 & 5000 & 0 & safe\\
			3 & 65 & 20 & 0 & 10 & 10 & 20 & 10000 & 1 & unsafe\\
			: &  &  &  &  &  &  &  &  & \\
			7535 & 65 & 30 & 0 & 10 & 10 & 20 & 2000 & 0 & safe\\
			7536 & 65 & 30 & 0 & 10 & 10 & 20 & 5000 & 1 & unsafe\\
			7537 & 65 & 30 & 0 & 10 & 10 & 20 & 10000 & 0 & safe\\
			: &  &  &  &  &  &  &  &  & \\
			16036 & 75 & 50 & 20 & 50 & 50 & 50 & 10000 & 1 & unsafe\\
			16037 & 75 & 50 & 20 & 50 & 50 & 100 & 2000 & 1 & unsafe\\
			16038 & 75 & 50 & 20 & 50 & 50 & 100 & 5000 & 1 & unsafe\\
			\noalign{\smallskip}\hline
		\end{tabular}
	\end{table}
	\begin{table}[b!]
		\caption{Samples of calibrated sensor responses based on the knowledge gathered from literature, data-sheets, lab tests, and scaling process.}
		\label{tab:1}       % Give a unique label
		\begin{tabular}{lllllllll}
			\hline\noalign{\smallskip}
			& & & \multicolumn{5}{c}{Sensors response ($\delta R_s/R_0$)} &  \\
			\cline{4-8}\noalign{\smallskip}
			\# &	Humidity &	Temperature &	InNO2 &	InCO &	InH2S &	InNH &	InCH &	Class \\
			\noalign{\smallskip}\hline\noalign{\smallskip}
			1 &	65 &	20 &	0.813 &	6.929 &	5.938 &	3.433 &	3.985 &	0 \\
			2 &	65 &	20 &	1.301 &	7.521 &	5.525 &	3.521 &	2.178 &	0 \\
			3 &	65 &	20 &	1.035 &	6.658 &	5.841 &	3.633 &	1.620 &	1 \\
			: &	   &  & & & & &  & \\						
			7535 &	65 &	30 &	1.038 &	7.565 &	5.658 &	3.228 &	2.275 &	0\\
			7536 &	65 &	30 &	1.054 &	6.694 &	5.745 &	3.692 &	1.268 &	1\\
			7537 &	65 &	30 &	0.642 &	7.210 &	5.819 &	1.326 &	3.530 &	0\\
			: &	   &  & & & & &  & \\							
			16036 &	75 &	50 &	4.645 &	2.764 &	0.608 &	2.709 &	0.499 &	1\\
			16037 &	75 &	50 &	4.712 &	2.985 &	0.641 &	1.228 &	0.450 &	1\\
			16038 &	75 &	50 &	4.911 &	2.433 &	0.381 &	0.937 &	0.481 &	1\\
			\noalign{\smallskip}\hline
		\end{tabular}
	\end{table}
	
	\subsection{Classification Based Approach}
	\label{sec:2.3}
	We categorized the classifiers in the four different groups of classifiers. Each category of classifiers contains three classifiers.
	\subsubsection{Network Based Classifiers}
	\paragraph{Multi-layer perceptron (MLP)} is a computational model that imitates human brain, and learn from environment, i.e., data. In our work, we used three-layered MLP, where layers are input layer, hidden layer, and output layer~\cite{weigend1990predicting}.
	\paragraph{Radial Basis Function Network (RBF)} is a special class of MLP, where inputs are mapped onto a hidden layer that consists of radial basis function, which does the non-linear mapping of input to a hidden layer~\cite{lowe2multi}.
	\paragraph{Support vector machine (SVM)} is a supervised learning computational model that maps input to a high dimension feature space using kernel trick. Hence, non-linear separable patterns in input space are linearly classified on a high dimensional feature space~\cite{cortes1995support}.  
	\subsubsection{Tree Based Classifiers}
	\paragraph{Reduced pruning tree (REP)} is a tree based classifier method, where a tree-like structure is designed for predicting target class based on the input variables~\cite{olshen1984classification,quinlan2014c4}. More specifically, the leaves of tree offers decision of the class based on the conjunction of the input feature represented by the branches of the tree. REP tree is a decision tree, where the tree size is reduced by pruning inefficient branches~\cite{esposito1999effects}.
	\paragraph{Naive Bayes tree (NBT)} is a special class of decision tree, where the leaf nodes of decision tree that offer decision on the class is replaced by a Naive Bayes classifier, which decides the class label, based on the features and learned threshold~\cite{mohamed2012comparative}.
	\paragraph{Logistic Model Trees (LMT)} is similar to NBT that does the transformation of leaves of a decision tree into a logistic regression node. A logistic regression maps independent variables to categorical dependent variables using a logistic function~\cite{walker1967estimation,cox1958regression}. Hence, LMT is a simple idea, where nodes of a decision/classification tree are replaced by logistic regression model~\cite{landwehr2005logistic}. 
	\subsubsection{Rule Based Classifiers}
	\paragraph{Decision Table (DT)} is a simple representation of data into a table based system, where the decision is made based on the features matching or searched into a decision table. On a successful search, the majority class label is returned, otherwise the majority class label of the entire dataset is returned as a decision for an unlabeled data~\cite{kohavi1995power}.
	\paragraph{PART} is a rule based classification method based on partial decision tree that generates a list of rules, used subsequently for making prediction of unknown data instance. The rules are generated based on the partial decision tree, which splits dataset into subsets until the entire dataset gets exhausted to form nodes and leaf nodes of the tree~\cite{frank1998generating}.
	\paragraph{Majority Predictor (Zero R)} is the simplest possible form of classification method. It is based on the majority of class label into a dataset. In simple words, it always predicts the majority class. 
	\subsubsection{Instance Based Classifiers}
	\paragraph{Instance-Based Learning (IBK)} provides the concept description which is the primary output of an IBK algorithm. It is a function that maps an instance to a category (class label). The concept description function is updated based on training procedure that involves two functions similarity and classification. The similarity function computes the similarity between the training instances and the pre-stored instances, and returns a numeric-value. Then, the classification function provides class label to the instances based on the results of similarity function. Accordingly, the concept description is updated~\cite{aha1991instance}.
	\paragraph{K$^ * $ (K Star)} is an instance-based learner that uses an entropy-based similarity matching function for searching/matching test instances to the learned instances~\cite{cleary1995k}.
	\paragraph{Locally Weighted Learning (LWL).} In a locally weighted learning, the prediction models are allowed to create at local points in a dataset or the specific point of interest rather than creating model for entire dataset. Hence, a linear regression or naive Bayes classifier or any other classifier may be used to create local models. In this case, we use Decision Stamp, which is a single level decision tree model for prediction~\cite{frank2002locally,atkeson1997lwl}.
	\subsubsection{Ensemble Methods}
	In this work, we tried to exploit different method of making ensemble. For an ensemble to perform well, we need to take into account two things which are accuracy of predictors and diversity among the predictors~\cite{polikar2006ensemble}. For example, Bagging maintains diversity by bootstrapping dataset, AddBoost combines several weak predictors, Random Subspace maintains diversity by splitting feature space, Random committee maintains diversity by creating predictors using different random seeds, and Rotation forest maintains diversity by splitting and extracting feature subspace using principal component analysis. Similarly, in multi-scheme and voting scheme, we combine several predictors to maintain diversity. Here, we describe the ensemble methods as follows. 
	
	\paragraph{Bagging.} In Bagging, several copies of same predictor is created. Each copy of the predictor learns a different replicate of learning set created from the complete training set using bootstrapping. Finally, the predictor's decision is combined using plurality voting method~\cite{breiman1996bagging}.
	\paragraph{Adaptive Boosting (AdaBoost)} is an ensemble technique that combines several weak predictors and inaccurate rules to create an accurate predictor~\cite{freund1997decision}. 
	\paragraph{Random Subspace (Random SUB).} In random subspace ensemble method feature space is divided into several feature subset. Hence, predictors are constructed for each feature subset. Finally, the decision of each constructed predictors are combined using voting method~\cite{ho1998random}.
	Random Committee (Random COM): In a random committee ensemble, several predictors are constructed over similar dataset, but they use different random seeds to maintain diversity in the ensemble. 
	\paragraph{Rotation Forest (Rotation FRST).} In this approach, training set for the predictors are created by splitting feature set into K subsets, and Principal Component Analysis is applied to extract all the principle components~\cite{rodriguez2006rotation}. Hence, diversity among the predictors are maintained by K axis rotation to form new feature set for training~\cite{rodriguez2006rotation}.
	\paragraph{Ensemble Selection (Ensemble SEL).} In the ensemble selection approach, the ensemble starts with an empty bag, and the predictors (chosen from a library of trained predictors) maximizing the performance of ensemble are added to the bag one by one to compute the decision of ensemble by using voting method~\cite{caruana2004ensemble}.
	\paragraph{Voting Scheme (Vote).} The voting scheme combines probability distribution of several chosen predictors/classifiers (or predictors available in a bag for making ensemble) using majority voting combination method~\cite{kuncheva2004combining}.
	\paragraph{Multi-Scheme (Multi).} The multi-scheme ensemble approach uses a bag of predictors and selects the output class by selecting a predictor from the bag of predictors based on cross-validation performance of the predictors~\cite{kuncheva2004combining}.
	\paragraph{Weighted Predictor Ensemble (WPE).} In this scheme of ensemble, the weight of predictors were determined. Subsequently, the ensemble output of $ k $ many predictors were computed as follows:
	$$  	y =  \arg \max_{j = 1}^c \sum\limits_{j = 1}^{k} w_j \mathbb{I}\left( P_j = \omega_j \right) ,  $$
	where $ c $ is the number of classes (here it is two), $ \mathbb{I}\left( P_j = \omega_j \right) $ is a function that returns value one for the predicted class $ \omega_j $.
	
	\section{Experimental Framework and Results}
	\label{sec:3}	
	Our aim in the experiment design was to obtain a highly accurate model for predicting hazardousness of the environment in a sewer pipeline. The sewer-pipeline environment was represented by the collected dataset. The second objective of the experiment design was to obtain results for analyzing the classifiers (predictors). Accordingly, the results of the classifiers were collected. Table~\ref{tab:2} represents the parameter setting of the chosen classifiers. For the evaluation of the classifiers, we repeated our experiments 10 times. Finally, the results were compared based on empirical and statistical (Kolmogorov--Smirnov test) evaluation. We used WEKA~\cite{Weka} and MATLAB tools~\cite{MATLAB} for the purpose of our experiments.
	
	\begin{table}
		\caption{Parameter setting of different classifiers}
		\label{tab:2}       % Give a unique label
		\begin{tabular}{llp{6.5cm}}
			\hline\noalign{\smallskip}
			category &  Classifiers &	 Parameters \\
			\noalign{\smallskip}\hline\noalign{\smallskip}
			\multirow{3}{2cm}{Network-Based Classifiers (F1)} &	MLP  &	Learning rate: 0.3, momentum factor: 0.2, iteration: 500, nodes in hidden layer: 100\\
			& RBF  &	Kernel: Gaussian basis function.\\
			& SVM  & 	Kernel: Radial basis function  \\
			\cline{2-3} 
			\multirow{3}{2cm}{Tree-Based Classifiers (F2)} & 	REP &  	Minimum no. of instance per leaf: 2, split proportion: 0.001\\
			& NBT & 	Leaf node: naïve Bayes classifier.\\
			& LMT & 	Node: logistic function, Number of instance per node for splitting: 15\\
			\cline{2-3} 
			\multirow{3}{2cm}{Instance-Based Classifiers (F3)} & 	IBK & 	Similarity function: linear nearest neighbor search, neighbor size: 1 \\
			& K Star & 	Similarity function: entropy distance measure.\\
			& LWL & 	Similarity function: linear nearest neighbor search, Weight function: Linear, Classifier: Decision Stamp.\\
			\cline{2-3} 
			\multirow{3}{2cm}{Rule-Based Classifiers (F4)} & 	DT &  	Evaluation metric: accuracy, Search method: best first\\
			& PART & 	Confidence threshold for pruning: 0.25\\
			& Zero R & 	-\\
			\cline{2-3} 
			\multirow{6}{2cm}{Ensemble Classifiers (E1)} & 	Bagging & 	Ensemble size: 10. Classifier: REP Tree \\
			& AdaBoost & 	Ensemble size: 10. Classifier: Decision Stamp.\\
			& Random SEL & Ensemble size: 10. Classifier:  REP Tree\\
			& Random COM & Ensemble size: 10. Classifier:  Random Tree\\
			& Rotation FRST & Ensemble size: 10. Classifier:  Random Tree\\
			& Ensemble SEL &  Ensemble size: 10. Classifier:  REP Tree\\
			\cline{2-3} 
			\multirow{3}{2cm}{Ensemble Classifiers (E2)} & 	Vote & 	Ensemble size: 12. Classifiers: F1, F2, F3 and F4 \\
			& Multi Scheme & 	Ensemble size: 12. Classifiers: F1, F2, F3 and F4\\
			& WPE & 	Ensemble size: 12. Classifiers: F1, F2, F3 and F4\\
			\noalign{\smallskip}\hline
		\end{tabular}
	\end{table}
	
	We organized the experimental results into three parts as reflected in Table 4. The first part in the table describes the category wise performance of classifier. Hence, the performance of the category of classifiers was evaluated. We represented the performance of the classifiers as per their training and test accuracy. An accuracy close to 1.0 indicates 100\% classification accuracy. Accordingly, the standard deviation (std) of training and test accuracies were reported for understanding the consistency of the classifiers' performance. In Table~\ref{tab:3}, the performance of the classifiers were arranged as follows. The category is arranged in the ascending order of their average accuracy over 10-fold CV test set, i.e., better performing classifier to the less performing classifier. The dataset was portioned into 10 equal sets and each time 9 sets were used for training and one set for testing. This process was repeated 10 times and each time a unique test set was used.  
	
	\begin{table}
		\caption{Experimental Results of Classifiers over 10 fold cross validation error}
		\label{tab:3}       % Give a unique label
		\begin{tabular}{llllll}
			\hline\noalign{\smallskip}
			category &   &	 \multicolumn{2}{c}{Training} & \multicolumn{2}{c}{Test} \\
			\cline{3-6}
			&  Classifiers  &	avg. accuracy  &	std  &	avg. accuracy  &	std \\
			\noalign{\smallskip}\hline\noalign{\smallskip}
			\multirow{3}{2cm}{NN-Based Classifiers} &
			SVM & 0.9407 & 0.0008 & 0.9340 & 0.0041\\
			& MLP & 0.8681 & 0.0029 & 0.8664 & 0.0081\\
			& RBF & 0.8064 & 0.0090 & 0.8051 & 0.0187\\
			\cline{2-6}\multirow{3}{2cm}{Tree-Based Classifiers} & 
			LMT & 0.9697 & 0.0039 & 0.9360 & 0.0023\\
			& REP Tree & 0.9528 & 0.0025 & 0.9265 & 0.0067\\
			& NB Tree & 0.9064 & 0.0469 & 0.8898 & 0.0418\\
			\cline{2-6}\multirow{3}{2cm}{Instance-Based Classifiers} & 
			IBK & 1.0000 & 0.0000 & 0.9671 & 0.0023\\
			& K Star & 0.9997 & 0.0001 & 0.9638 & 0.0036\\
			& LWL & 0.7613 & 0.0014 & 0.7613 & 0.0128\\
			\cline{2-6}\multirow{3}{2cm}{Rule-Based Classifiers} & 
			PART & 0.9275 & 0.0097 & 0.9062 & 0.0103\\
			& Decision Table & 0.8672 & 0.0049 & 0.8553 & 0.0154\\
			& Zero R & 0.7613 & 0.0010 & 0.7613 & 0.0091\\
			\cline{2-6}\multirow{9}{2cm}{Ensemble Classifiers} & 
			Multi & 1.0000 & 0.0000 & 0.9672 & 0.0035\\
			& Rotation FRST & 1.0000 & 0.0000 & 0.9622 & 0.0036\\
			& Random COM & 1.0000 & 0.0000 & 0.9549 & 0.0073\\
			& Bagging & 0.9728 & 0.0008 & 0.9395 & 0.0077\\
			& WPE & 0.9635 & 0.0006 & 0.9356 & 0.0056\\
			& Ensemble SEL & 0.9577 & 0.0009 & 0.9330 & 0.0077\\
			& Vote & 0.9423 & 0.0128 & 0.9214 & 0.0143\\
			& Random SUB & 0.9160 & 0.0103 & 0.8720 & 0.0165\\
			& AdaBoostM1 & 0.7613 & 0.0010 & 0.7613 & 0.0091\\
			\noalign{\smallskip}\hline
		\end{tabular}
	\end{table}
	In the second part, we organized the results according to rank of the classifiers' performance over 10-fold test set. It may please be noted that for each classifier, we collected 10 instances of 10-fold CV training and test results. Hence, the results in Table~\ref{tab:4} reflect averaged training and test accuracy of the classifiers. However, ranking the classifiers based only on the average results does not say much about the quality of the classifier. Hence, in the third part of the results, we used pairwise comparison of the classifiers using Kolmogorov--Smirnov (KS) test, which ascertains whether the supremacy  of one classifier over the other is statistically significant or not. A comprehensive matrix of the pairwise KS test results are presented in Table~\ref{tab:5}.  The KS Test is a non-parametric statistical test that determines the difference between the cumulative frequency distribution (cfd) of two samples. In other words, it indicates  whether the empirical cfd of one sample is equal ``=", larger ``$\succ$", or smaller ``$\prec$" than the other. It tells whether two dataset A and B are statistically similar ``A=B", dissimilar ``A$\prec$B", where A being statistically dominated by B, or dissimilar ``A$\succ$B", where A being statistically dominant over B. In our experiments, the KS test was evaluated with 5\% significance level, i.e., with 95\% confidence. 
	
	\begin{table}
		\caption{Ranking algorithms according to their performance on test set (10 Fold CV).}
		\label{tab:4}       % Give a unique label
		\begin{tabular}{lllll}
			\hline\noalign{\smallskip}
			Rank & category & Classifiers & Training & Test\\
			\noalign{\smallskip}\hline\noalign{\smallskip}
			1 & E2 & Multi & 100.0000 & 96.8060\\
			2 & F3 & IBK & 100.0000 & 96.7945\\
			3 & F3 & KStar & 99.9653 & 96.4677\\
			4 & E1 & Rotation FRST & 100.0000 & 96.2725\\
			5 & E1 & Random COM & 100.0000 & 95.7737\\
			6 & E1 & Bagging & 97.2874 & 94.1025\\
			7 & E2 & WPE & 96.3564 & 93.9865\\
			8 & F2 & LMT & 96.7454 & 93.4674\\
			9 & E1 & Ensemble SEL & 95.7173 & 93.3778\\
			10 & F1 & SVM & 94.0837 & 93.3647\\
			11 & F2 & REPTree & 95.2469 & 92.4733\\
			12 & E2 & Vote & 93.9593 & 92.1314\\
			13 & F4 & PART & 92.7331 & 90.8675\\
			14 & F2 & NBTree & 89.4770 & 88.0795\\
			15 & F1 & MLP & 86.8566 & 86.6336\\
			16 & E1 & Random SUB & 90.9650 & 86.4397\\
			17 & F4 & DT & 86.6874 & 85.4790\\
			18 & F1 & RBF & 80.7795 & 80.7571\\
			19 & F3 & LWL & 76.1285 & 76.1451\\
			20 & F4 & ZeroR & 76.1285 & 76.1451\\
			21 & E1 & AdaBoost & 76.1302 & 76.1302\\
			\noalign{\smallskip}\hline
		\end{tabular}
	\end{table}
	
	\begin{table}
		\caption{Ranking algorithms according to their performance on test set (10 Fold CV).}
		\label{tab:5}       % Give a unique label
		\setlength{\tabcolsep}{0.06cm}
		\begin{tabular}{l|lllllllllllllllllllll}
			\hline\noalign{\smallskip}
			Classifiers & \begin{sideways} \makecell[l]{DT}\end{sideways} & \begin{sideways} \makecell[l]{IBK}\end{sideways} & \begin{sideways} \makecell[l]{Kstar}\end{sideways} & \begin{sideways} \makecell[l]{LMT}\end{sideways} & \begin{sideways} \makecell[l]{LWL}\end{sideways} &\begin{sideways} \makecell[l]{ MLP}\end{sideways} &\begin{sideways} \makecell[l]{NBTree}\end{sideways} & \begin{sideways} \makecell[l]{PART}\end{sideways} & \begin{sideways} \makecell[l]{RBF}\end{sideways} & \begin{sideways} \makecell[l]{REPTree}\end{sideways} & \begin{sideways} \makecell[l]{SVM}\end{sideways} & \begin{sideways} \makecell[l]{ZeroR}\end{sideways} & \begin{sideways} \makecell[l]{Multi-Scheme}\end{sideways} & \begin{sideways} \makecell[l]{Vote}\end{sideways} & \begin{sideways} \makecell[l]{AdaBoost}\end{sideways} & \begin{sideways} \makecell[l]{Bagging}\end{sideways} & \begin{sideways} \makecell[l]{Ensemble SEL}\end{sideways} & \begin{sideways} \makecell[l]{Random COM}\end{sideways} & \begin{sideways} \makecell[l]{Random SUB}\end{sideways} & \begin{sideways} \makecell[l]{Rotation FRST}\end{sideways} & \begin{sideways} \makecell[l]{WPE}\end{sideways}\\
			\noalign{\smallskip}\hline\noalign{\smallskip}
			DT & . & $\prec$ & $\prec$ & $\prec$ & $\succ$ & $\prec$ & $\prec$ & $\prec$ & $\succ$ & $\prec$ & $\prec$ & $\succ$ & $\prec$ & $\prec$ & $\succ$ & $\prec$ & $\prec$ & $\prec$ & $\prec$ & $\prec$ & $\prec$\\
			IBK & . & . & $\succ$ & $\succ$ & $\succ$ & $\succ$ & $\succ$ & $\succ$ & $\succ$ & $\succ$ & $\succ$ & $\succ$ & $\succ$ & $\succ$ & $\succ$ & $\succ$ & $\succ$ & $\succ$ & $\succ$ & $\succ$ & $\succ$\\
			Kstar & . & . & . & $\succ$ & $\succ$ & $\succ$ & $\succ$ & $\succ$ & $\succ$ & $\succ$ & $\succ$ & $\succ$ & $\prec$ & $\succ$ & $\succ$ & $\succ$ & $\succ$ & $\succ$ & $\succ$ & $\succ$ & $\succ$\\
			LMT & . & . & . & . & $\succ$ & $\succ$ & $\succ$ & $\succ$ & $\succ$ & $\succ$ & $\succ$ & $\succ$ & $\prec$ & $\succ$ & $\succ$ & $\prec$ & $\succ$ & $\prec$ & $\succ$ & $\prec$ & $\prec$\\
			LWL & . & . & . & . & . & $\prec$ & $\prec$ & $\prec$ & $\prec$ & $\prec$ & $\prec$ & = & $\prec$ & $\prec$ & = & $\prec$ & $\prec$ & $\prec$ & $\prec$ & $\prec$ & $\prec$\\
			MLP & . & . & . & . & . & . & $\prec$ & $\prec$ & $\succ$ & $\prec$ & $\prec$ & $\succ$ & $\prec$ & $\prec$ & $\succ$ & $\prec$ & $\prec$ & $\prec$ & = & $\prec$ & $\prec$\\
			NBTree & . & . & . & . & . & . & . & $\prec$ & $\succ$ & $\prec$ & $\prec$ & $\succ$ & $\prec$ & $\prec$ & $\succ$ & $\prec$ & $\prec$ & $\prec$ & $\succ$ & $\prec$ & $\prec$\\
			PART & . & . & . & . & . & . & . & . & $\succ$ & $\prec$ & $\prec$ & $\succ$ & $\prec$ & $\prec$ & $\succ$ & $\prec$ & $\prec$ & $\prec$ & $\succ$ & $\prec$ & $\prec$\\
			RBF & . & . & . & . & . & . & . & . & . & $\prec$ & $\prec$ & $\succ$ & $\prec$ & $\prec$ & $\succ$ & $\prec$ & $\prec$ & $\prec$ & $\prec$ & $\prec$ & $\prec$\\
			REPTree & . & . & . & . & . & . & . & . & . & . & $\prec$ & $\succ$ & $\prec$ & $\succ$ & $\succ$ & $\prec$ & $\prec$ & $\prec$ & $\succ$ & $\prec$ & $\prec$\\
			SVM & . & . & . & . & . & . & . & . & . & . & . & $\succ$ & $\prec$ & $\succ$ & $\succ$ & $\prec$ & = & $\prec$ & $\succ$ & $\prec$ & $\prec$\\
			ZeroR & . & . & . & . & . & . & . & . & . & . & . & . & $\prec$ & $\prec$ & = & $\prec$ & $\prec$ & $\prec$ & $\prec$ & $\prec$ & $\succ$\\
			Multi-Scheme & . & . & . & . & . & . & . & . & . & . & . & . & . & $\succ$ & $\succ$ & $\succ$ & $\succ$ & $\succ$ & $\succ$ & $\succ$ & $\prec$\\
			Vote & . & . & . & . & . & . & . & . & . & . & . & . & . & . & $\succ$ & $\prec$ & $\prec$ & $\prec$ & $\succ$ & $\prec$ & $\prec$\\
			AdaBoost & . & . & . & . & . & . & . & . & . & . & . & . & . & . & . & $\prec$ & $\prec$ & $\prec$ & $\prec$ & $\prec$ & $\prec$\\
			Bagging & . & . & . & . & . & . & . & . & . & . & . & . & . & . & . & . & $\succ$ & $\prec$ & $\succ$ & $\prec$ & $\succ$\\
			Ensemble SEL & . & . & . & . & . & . & . & . & . & . & . & . & . & . & . & . & . & $\prec$ & $\succ$ & $\prec$ & $\succ$\\
			Random COM & . & . & . & . & . & . & . & . & . & . & . & . & . & . & . & . & . & . & $\succ$ & $\prec$ & $\succ$\\
			Random SUB & . & . & . & . & . & . & . & . & . & . & . & . & . & . & . & . & . & . & . & $\prec$ & $\prec$\\
			Rotation FRST & . & . & . & . & . & . & . & . & . & . & . & . & . & . & . & . & . & . & . & . & $\succ$\\
			WPE & . & . & . & . & . & . & . & . & . & . & . & . & . & . & . & . & . & . & . & . & .\\
			\noalign{\smallskip}\hline
		\end{tabular}
	\end{table}

	\section{Discussions}
	\label{sec:4}
	Since the developed electronic portable gas detector shall be used by naive persons who are engaged in maintaining sewer-pipeline, we are looking for binary answer. Hence, our objective is to search for classification accuracy and the model (weights) with the highest accuracy so that such a combination may be implemented into electronic portable gas detector form. 
	
	Moreover, it is also a difficult task to be certain with the accuracy of an implemented electronic portable gas detector because the toxic exposure of a gas is also proportional to the time and not only its safety limit. However, with a real-time monitoring and requisite maintenance involved, the accuracy of detector may be relaxed and hence, we resorted to choose 90\% accuracy as the accuracy for our developed detector. So, the classifier's performance was compared with a threshold setting of 90\% accuracy. 
	
	First, let us discuss on  the obtained results. For  the classifiers belonging to network-based category F1, the classifier SVM performs better than its counterparts MLP and RBF both in terms of  high accuracy (test accuracy 0.93403) and high consistency (std on test accuracy 0.0041). On the other hand, the performance of MLP was reported next to SVM with high consistency. The performance of the RBF was found to be inconsistent and poorer in comparison to its counterparts. 
	
	In the tree based category F2, the performance of LMT and REPTree was comparable to whereas, NBTree has shown poor performance compared to its counterparts. 
	
	In instance-base category F3, the performance of IBK and K Star was comparative with a high accuracy and high consistency. LWL performed poor with a very low accuracy. 
	
	When it came to the category of rule based classifier F4, PART has outperformed others in its category, but the consistency was not as high as the consistency of the other well performing classifiers IBK, SVM, MLP, etc. The classifier ZeroR consistently performed poor in comparison to all other classifiers. 
	
	In the ensemble category E1 and E2, the multi-scheme, Random COM, Rotation FRST, Bagging, WPE, and Ensemble SEL performed with high accuracies (over 90\%) and consistency. However, the performance of the ensembles Random Forest, Vote, and AdaBoost were not as satisfactory as compared to the other ensembles. One of the reason behind poor performance of Random SUB was the usage of subset of the features. Therefore, the feature selection may not help in case of this dataset because of the high correlation maintained by each of the features with the output feature. Similarly, Voting used probability measures to combine the predictors and AddBoost combined weak predictors, whereas, the entirely better performing ensemble exploited the best predictors. Hence, they performed better in this scenario.  
	
	Considering the assumption of 90\% accuracy being a good predictor for implementation as gas detector, we can figure out from  Table~\ref{tab:4} that the classifiers belong to  category F3 (exception of the classifier LWL) had performed better than the classifiers of other categories. However, the instance based classifier IBK is not suitable for the implementation as electronic gas detector since it required a large memory for its computation for saving all the instances of the training set. IBK prediction is computed based on all training samples. Hence, it takes long time to compute the output, which is unacceptable  in real time. 
	
	The next category whose performance was found close to IBK were the classifiers of category F2 (tree based classifier). Two classifies, LMT and REP Tree qualified the 90\% accuracy threshold. On the contrary, two classifiers from each F1 and F4 had performed lower than 90\% accuracy. However, SVM performed significantly well with a very high accuracy 93.36\%. Similarly, classifier PART from category F4 had an accuracy of 90.86\%. However, since the SVM produced less number of parameters than the tree based predictor and it robustly accommodates the noisy attributes, it was recommended from these experiments that SVM is a proper choice for the implementation of the proposed gas detector.      
	
	\section{Conclusion}
	\label{sec:5}
	In this work, we explored a real world problem in the context of classification, where we simplified the approach by offering binary decision to the problem. We explored the problem related to the detection of hazardousness of a sewer pipeline environment. This is very crucial problem since it is related to the safety of the persons who have to work under the toxic environment of the sewer-pipeline. Usually, a sewer-pipeline environment contains mixture of toxic gases. Hence, we collected samples from sewer pipelines from different locations. Then we examined those samples to identify data samples for our experiments. We prepared a large dataset by collecting gas sensor responses from laboratory tests, literature and scaled the collected gas sensor responses to form a dataset where non-hazardous samples were labeled 0 and hazardous samples were labeled 1. Finally, we applied 21 different classifiers over the identified dataset and their empirical and statistical performance were evaluated. We discovered that for this problem, the instance based classifier performed best followed by the performance of tree based classifiers. However, we found that the performance of the classifiers were dependent on the ability and mechanism of the classifiers themselves and not on the information regarding which category they belong to.

	\begin{acknowledgements}
		This work was supported by the IPROCOM Marie Curie Initial Training Network, funded through the People Programme (Marie Curie Actions) of the European Union’s Seventh Framework Programme.
	\end{acknowledgements}
	
	\bibliographystyle{IEEEtran}
	%\bibliography{lit_varun_ju,lit_algo_sa,lit_sewer}
	%5\bibliographystyle{spbasic}      % basic style, author-year citations
	\bibliography{class_nca_rev}   % name your BibTeX data base

\begin{thebibliography}{10}
\providecommand{\url}[1]{#1}
\csname url@samestyle\endcsname
\providecommand{\newblock}{\relax}
\providecommand{\bibinfo}[2]{#2}
\providecommand{\BIBentrySTDinterwordspacing}{\spaceskip=0pt\relax}
\providecommand{\BIBentryALTinterwordstretchfactor}{4}
\providecommand{\BIBentryALTinterwordspacing}{\spaceskip=\fontdimen2\font plus
\BIBentryALTinterwordstretchfactor\fontdimen3\font minus
  \fontdimen4\font\relax}
\providecommand{\BIBforeignlanguage}[2]{{%
\expandafter\ifx\csname l@#1\endcsname\relax
\typeout{** WARNING: IEEEtran.bst: No hyphenation pattern has been}%
\typeout{** loaded for the language `#1'. Using the pattern for}%
\typeout{** the default language instead.}%
\else
\language=\csname l@#1\endcsname
\fi
#2}}
\providecommand{\BIBdecl}{\relax}
\BIBdecl

\bibitem{SewerGas}
J.~Whorton, ````the insidious foe"--sewer gas,'' \emph{Western Journal of
  Medicine}, vol. 175, no.~6, pp. 427--–428, Dec. 2001.

\bibitem{lewis1996material}
R.~J. Lewis, \emph{Sax's Dangerous Properties of Industrial Materials},
  12th~ed.\hskip 1em plus 0.5em minus 0.4em\relax Wiley, 2010.

\bibitem{SewerGas1}
N.~Gromicko, ``Sewer gases in the home,'' 2006,
  http://www.nachi.org/sewer-gases-home.html.

\bibitem{Hindu2014a}
T.~Hindu, ``Deaths in the drains,'' 2014,
  http://www.thehindu.com/opinion/op-ed/deaths-in-the-drains/article5868090.ece?homepage=true.,
  Accessed on 15 Dec 2015.

\bibitem{NDTV2015}
NDTV, ``He died on diwali inside a sewage pipe,'' 2014,
  http://www.ndtv.com/opinion/he-died-on-diwali-inside-a-sewage-pipe-1245559,
  Accessed on 15 Dec 2015.

\bibitem{Tehelka2007}
S.~Anand, ``Dying in the gutters,'' \emph{Tehelka Magazine}, vol.~4, no.~47,
  Dec 2007,
  http://archive.tehelka.com/story\_main36.asp?filename=Ne081207DYING.asp,
  Accessed on: 15 Dec 2015.

\bibitem{Hindu2014d}
T.~Hindu, ``Provide safety gear to sewer workers who enter manholes, says
  court,'' 2011,
  http://www.thehindu.com/todays-paper/tp-national/provide-safety-gear-to-sewer-workers-who-enter-manholes-says-court/article2228688.ece,
  Accessed on 15 Dec 2015.

\bibitem{Hindu2014b}
------, ``Sewer deaths,'' 2014,
  http://www.thehindu.com/opinion/letters/sewer-deaths/ article5873493.ece,
  Accessed on 15 Dec 2015.

\bibitem{Hindu2014c}
------, ``Supreme court orders states to abolish manual scavenging,'' 2014,
  http://www.thehindu.com/news/national/supreme-court-orders-states-to-abolish-manual-scavenging/article5840086.ece,
  Accessed on 15 Dec 2015.

\bibitem{wolpert1997no}
D.~H. Wolpert and W.~G. Macready, ``No free lunch theorems for optimization,''
  \emph{IEEE Transactions on Evolutionary Computation}, vol.~1, no.~1, pp.
  67--82, 1997.

\bibitem{jun1993enose}
J.~Li, ``A mixed gas sensor system based on thin film saw sensor array and
  neural network,'' in \emph{Proceedings of the Twelfth Southern Biomedical
  Engineering Conference}, 1993, pp. 179--181.

\bibitem{sirvastava2000enose}
A.~Srivastava, S.~Srivastava, and K.~Shukla, ``On the design issue of
  intelligent electronic nose system,'' in \emph{Proceedings of IEEE
  International Conference on Industrial Technology 2000.}, vol.~2.\hskip 1em
  plus 0.5em minus 0.4em\relax IEEE, 2000, pp. 243--248.

\bibitem{sirvastava2000enoseGA}
------, ``In search of a good neuro-genetic computational paradigm,'' in
  \emph{Proceedings of IEEE International Conference on Industrial Technology
  2000.}, vol.~1.\hskip 1em plus 0.5em minus 0.4em\relax IEEE, 2000, pp.
  497--502.

\bibitem{eduard2001enose}
E.~Llobet, R.~Ionescu, S.~Al-Khalifa, J.~Brezmes, X.~Vilanova, X.~Correig,
  N.~Barsan, and J.~W. Gardner, ``Multicomponent gas mixture analysis using a
  single tin oxide sensor and dynamic pattern recognition,'' \emph{IEEE Sensors
  Journal}, vol.~1, no.~3, pp. 207--213, 2001.

\bibitem{dae2005enose}
D.-S. Lee, S.-W. Ban, M.~Lee, and D.-D. Lee, ``Micro gas sensor array with
  neural network for recognizing combustible leakage gases,'' \emph{IEEE
  Sensors Journal}, vol.~5, no.~3, pp. 530--536, 2005.

\bibitem{maxim2008enose}
M.~Ambard, B.~Guo, D.~Martinez, and A.~Bermak, ``A spiking neural network for
  gas discrimination using a tin oxide sensor array,'' in \emph{4th IEEE
  International Symposium on Electronic Design, Test and Applications}.\hskip
  1em plus 0.5em minus 0.4em\relax IEEE, 2008, pp. 394--397.

\bibitem{hakim2009enose}
H.~Baha and Z.~Dibi, ``A novel neural network-based technique for smart gas
  sensors operating in a dynamic environment,'' \emph{Sensors}, vol.~9, no.~11,
  pp. 8944--8960, 2009.

\bibitem{wu2009enose}
W.~Pan, N.~Li, and P.~Liu, ``Application of electronic nose in gas mixture
  quantitative detection,'' in \emph{IEEE International Conference on Network
  Infrastructure and Digital Content.}\hskip 1em plus 0.5em minus 0.4em\relax
  IEEE, 2009, pp. 976--980.

\bibitem{chatchawal2010enose}
C.~Wongchoosuk, A.~Wisitsoraat, A.~Tuantranont, and T.~Kerdcharoen, ``Portable
  electronic nose based on carbon nanotube- {SnO}$_2$ gas sensors and its
  application for detection of methanol contamination in whiskeys,''
  \emph{Sensors and Actuators B: Chemical}, vol. 147, no.~2, pp. 392--399,
  2010.

\bibitem{qian2010enose}
Q.~Zhang, H.~Li, and Z.~Tang, ``Knowledge-based genetic algorithms data fusion
  and its application in mine mixed-gas detection,'' in \emph{Chinese Control
  and Decision Conference (CCDC)}.\hskip 1em plus 0.5em minus 0.4em\relax IEEE,
  2010, pp. 1334--1338.

\bibitem{won2010enose}
W.~So, J.~Koo, D.~Shin, and E.~S. Yoon, ``The estimation of hazardous gas
  release rate using optical sensor and neural network,'' \emph{Computer Aided
  Chemical Engineering}, vol.~28, pp. 199--204, 2010.

\bibitem{Ojha_ga_j}
V.~K. Ojha, P.~Dutta, and H.~Saha, ``Performance analysis of neuro genetic
  algorithm applied on detecting proportion of components in manhole gas
  mixture,'' \emph{International Journal of Artificial Intelligence
  $\backslash$\& Applications}, vol.~3, no.~4, pp. 83--98, 2012.

\bibitem{Ojha_pso_j}
V.~K. Ojha and P.~Dutta, ``Performance analysis of neuro swarm optimization
  algorithm applied on detecting proportion of components in manhole gas
  mixture,'' \emph{Artificial Intelligence Research}, vol.~1, no.~1, pp.
  31--45, 2012.

\bibitem{Ojha_ch_bp}
V.~K. Ojha, P.~Dutta, A.~Chaudhuri, and H.~Saha, ``Convergence analysis of
  backpropagation algorithm for designing an intelligent system for sensing
  manhole gases,'' in \emph{Hybrid Soft Computing Approaches}.\hskip 1em plus
  0.5em minus 0.4em\relax Springer India, 2016, pp. 215--236.

\bibitem{Ojha_ch_cg}
P.~Dutta and V.~K. Ojha, ``Conjugate gradient trained neural network for
  intelligent sensing of manhole gases to avoid human fatality,'' in
  \emph{Advances in Secure Computing, Internet Services, and
  Applications}.\hskip 1em plus 0.5em minus 0.4em\relax IGI Global, 2013, pp.
  257--280.

\bibitem{ojha2016understating}
V.~K. Ojha, P.~Dutta, A.~Chaudhuri, and H.~Saha, ``Understating continuous ant
  colony optimization for neural network training: A case study on intelligent
  sensing of manhole gas components,'' \emph{International Journal of Hybrid
  Intelligent Systems}, vol.~12, no.~4, pp. 185--202, 2016.

\bibitem{ojha2016multi}
------, ``A multi-agent concurrent neurosimulated annealing algorithm: A case
  study on intelligent sensing of manhole gases,'' \emph{International Journal
  of Hybrid Intelligent Systems}, vol.~12, no.~4, pp. 203--217, 2016.

\bibitem{GSA}
S.~Ghosh, A.~Roy, S.~Singh, H.~Saha, V.~K. Ojha, and P.~Dutta, ``Sensor array
  for manhole gas analysis,'' in \emph{1st International Symposium on Physics
  and Technology of Sensors (ISPTS)}.\hskip 1em plus 0.5em minus 0.4em\relax
  IEEE, 2012, pp. 9--12.

\bibitem{PortableGSA}
S.~Ghosh, H.~Saha, C.~RoyChaudhuri, V.~K. Ojha, and P.~Dutta, ``Portable sensor
  array system for intelligent recognizer of manhole gas,'' in \emph{Sixth
  International Conference on Sensing Technology (ICST)}.\hskip 1em plus 0.5em
  minus 0.4em\relax IEEE, 2012, pp. 589--594.

\bibitem{cantalini2003sensitivity}
C.~Cantalini, L.~Valentini, I.~Armentano, L.~Lozzi, J.~Kenny, and S.~Santucci,
  ``Sensitivity to {NO}$_2$ and cross-sensitivity analysis to {NH}$_3$, ethanol
  and humidity of carbon nanotubes thin film prepared by {PECVD},''
  \emph{Sensors and Actuators B: Chemical}, vol.~95, no.~1, pp. 195--202, 2003.

\bibitem{mitzner2003development}
K.~D. Mitzner, J.~Sternhagen, and D.~W. Galipeau, ``Development of a
  micromachined hazardous gas sensor array,'' \emph{Sensors and Actuators B:
  Chemical}, vol.~93, no.~1, pp. 92--99, 2003.

\bibitem{junhua2001enose}
J.~Liu, Y.~Zhang, Y.~Zhang, and M.~Cheng, ``Cross sensitivity reduction of gas
  sensors using genetic algorithm neural network,'' in \emph{Optical Methods
  for Industrial Processes}, S.~Farquharson, Ed., vol. 4201.\hskip 1em plus
  0.5em minus 0.4em\relax Proceedings of SPIE, 2001.

\bibitem{donham2002exposure}
K.~J. Donham, ``Exposure limits related to air quality and risk assessment,''
  \emph{Iowa Concentrated Animal Feeding Operations Air Quality Study}, p. 164,
  2002.

\bibitem{gasCO1998}
L.~K. Weaver, ``Carbon monoxide poisoning,'' \emph{New England Journal of
  Medicine}, vol. 360, no.~12, pp. 1217--1225, 2009.

\bibitem{gasH2S2007}
S.~Simonton, ``Human health effects from exposure to low-level concentrations
  of hydrogen sulfide,'' \emph{Occupational Health \& Safety}, Nov. 2007.

\bibitem{gasCO22007}
G.~Shilpa, ``New insight into panic attacks: Carbon dioxide is the culprit,''
  \emph{Journal of Young Investigators}, Nov. 2007,
  http://www.jyi.org/issue/new-insight-into-panic-attacks-carbon-dioxide-is-the-culprit/.

\bibitem{gasCH42002}
D.~W. Fahey and M.~I. Hegglin, ``Twenty questions and answers about the ozone
  layer: 2010 update,'' \emph{Scientific assessment of ozone depletion}, pp.
  4--1, 2010.

\bibitem{weigend1990predicting}
A.~S. Weigend, B.~A. Huberman, and D.~E. Rumelhart, ``Predicting the future: A
  connectionist approach,'' \emph{International journal of neural systems},
  vol.~1, no.~03, pp. 193--209, 1990.

\bibitem{lowe2multi}
D.~Lowe and D.~Broomhead, ``Multivariable functional interpolation and adaptive
  networks,'' \emph{Complex System}, vol.~2, pp. 321--355, 1988.

\bibitem{cortes1995support}
C.~Cortes and V.~Vapnik, ``Support-vector networks,'' \emph{Machine learning},
  vol.~20, no.~3, pp. 273--297, 1995.

\bibitem{olshen1984classification}
L.~Olshen, C.~J. Stone \emph{et~al.}, ``Classification and regression trees,''
  \emph{Wadsworth International Group}, vol.~93, no.~99, p. 101, 1984.

\bibitem{quinlan2014c4}
J.~R. Quinlan, \emph{C4. 5: programs for machine learning}.\hskip 1em plus
  0.5em minus 0.4em\relax Elsevier, 2014.

\bibitem{esposito1999effects}
F.~Esposito, D.~Malerba, G.~Semeraro, and V.~Tamma, ``The effects of pruning
  methods on the predictive accuracy of induced decision trees,'' \emph{Applied
  Stochastic Models in Business and Industry}, vol.~15, no.~4, pp. 277--299,
  1999.

\bibitem{mohamed2012comparative}
W.~N. H.~W. Mohamed, M.~N.~M. Salleh, and A.~H. Omar, ``A comparative study of
  reduced error pruning method in decision tree algorithms,'' in \emph{IEEE
  International Conference on Control System, Computing and Engineering
  (ICCSCE), 2012}.\hskip 1em plus 0.5em minus 0.4em\relax IEEE, 2012, pp.
  392--397.

\bibitem{walker1967estimation}
S.~H. Walker and D.~B. Duncan, ``Estimation of the probability of an event as a
  function of several independent variables,'' \emph{Biometrika}, vol.~54, no.
  1-2, pp. 167--179, 1967.

\bibitem{cox1958regression}
D.~R. Cox, ``The regression analysis of binary sequences,'' \emph{Journal of
  the Royal Statistical Society. Series B (Methodological)}, pp. 215--242,
  1958.

\bibitem{landwehr2005logistic}
N.~Landwehr, M.~Hall, and E.~Frank, ``Logistic model trees,'' \emph{Machine
  Learning}, vol.~59, no. 1-2, pp. 161--205, 2005.

\bibitem{kohavi1995power}
R.~Kohavi, ``The power of decision tables,'' in \emph{Machine Learning:
  ECML-95}.\hskip 1em plus 0.5em minus 0.4em\relax Springer, 1995, pp.
  174--189.

\bibitem{frank1998generating}
E.~Frank and I.~H. Witten, ``Generating accurate rule sets without global
  optimization,'' 1998.

\bibitem{aha1991instance}
D.~W. Aha, D.~Kibler, and M.~K. Albert, ``Instance-based learning algorithms,''
  \emph{Machine learning}, vol.~6, no.~1, pp. 37--66, 1991.

\bibitem{cleary1995k}
J.~G. Cleary, L.~E. Trigg \emph{et~al.}, ``K*: An instance-based learner using
  an entropic distance measure,'' in \emph{Proceedings of the 12th
  International Conference on Machine learning}, vol.~5, 1995, pp. 108--114.

\bibitem{frank2002locally}
E.~Frank, M.~Hall, and B.~Pfahringer, ``Locally weighted naive bayes,'' in
  \emph{Proceedings of the Nineteenth conference on Uncertainty in Artificial
  Intelligence}.\hskip 1em plus 0.5em minus 0.4em\relax Morgan Kaufmann
  Publishers Inc., 2002, pp. 249--256.

\bibitem{atkeson1997lwl}
C.~G. Atkeson, A.~W. Moore, and S.~Schaal, ``Locally weighted learning,''
  \emph{Artificial Intelligence Review}, vol.~11, no.~5, pp. 11--73, 1997.

\bibitem{polikar2006ensemble}
R.~Polikar, ``Ensemble based systems in decision making,'' \emph{IEEE Circuits
  and Systems Magazine}, vol.~6, no.~3, pp. 21--45, 2006.

\bibitem{breiman1996bagging}
L.~Breiman, ``Bagging predictors,'' \emph{Machine learning}, vol.~24, no.~2,
  pp. 123--140, 1996.

\bibitem{freund1997decision}
Y.~Freund and R.~E. Schapire, ``A decision-theoretic generalization of on-line
  learning and an application to boosting,'' \emph{Journal of computer and
  system sciences}, vol.~55, no.~1, pp. 119--139, 1997.

\bibitem{ho1998random}
T.~K. Ho, ``The random subspace method for constructing decision forests,''
  \emph{IEEE Transactions on Pattern Analysis and Machine Intelligence},
  vol.~20, no.~8, pp. 832--844, 1998.

\bibitem{rodriguez2006rotation}
J.~J. Rodriguez, L.~I. Kuncheva, and C.~J. Alonso, ``Rotation forest: A new
  classifier ensemble method,'' \emph{IEEE Transactions on Pattern Analysis and
  Machine Intelligence}, vol.~28, no.~10, pp. 1619--1630, 2006.

\bibitem{caruana2004ensemble}
R.~Caruana, A.~Niculescu-Mizil, G.~Crew, and A.~Ksikes, ``Ensemble selection
  from libraries of models,'' in \emph{Proceedings of the twenty-first
  international conference on Machine learning}.\hskip 1em plus 0.5em minus
  0.4em\relax ACM, 2004, p.~18.

\bibitem{kuncheva2004combining}
L.~I. Kuncheva, \emph{Combining pattern classifiers: methods and
  algorithms}.\hskip 1em plus 0.5em minus 0.4em\relax John Wiley \& Sons, 2004.

\bibitem{Weka}
\BIBentryALTinterwordspacing
``Weka 3: Data mining software in java,'' accessed: 2016-05-01. [Online].
  Available: \url{http://www.cs.waikato.ac.nz/ml/index.html}
\BIBentrySTDinterwordspacing

\bibitem{MATLAB}
\BIBentryALTinterwordspacing
``Matlab: Statistics and machine learning toolbox,'' accessed: 2016-05-01.
  [Online]. Available: \url{http://www.mathworks.com/products/matlab/}
\BIBentrySTDinterwordspacing

\end{thebibliography}
	
	%\begin{thebibliography}{99}
	%\bibitem{key}{text}.
	%\end{thebibliography}
	
\end{document}